\documentclass[runningheads]{llncs}
\usepackage{graphicx}

\usepackage{tikz}
\usepackage{comment}
\usepackage{amsmath,amssymb} %
\usepackage{color}

\usepackage[accsupp]{axessibility}  %

\usepackage{booktabs}
\usepackage{epsfig}
\usepackage{subcaption} %

\usepackage[font=small]{caption}
\usepackage{multirow}
\usepackage{multicol}
\usepackage{arydshln}
\usepackage{capt-of}
\usepackage{algorithm}
\usepackage{listings}
\usepackage{tabularx}
\usepackage{colortbl}
\usepackage{cuted}
\usepackage{adjustbox}
\usepackage{enumitem}
\usepackage{wrapfig}
\usepackage{amsmath}
\usepackage{microtype}
\usepackage{xcolor}

\definecolor{newlightblue}{RGB}{0,75,255}
\usepackage[pagebackref,breaklinks,colorlinks, urlcolor={newlightblue}, citecolor={newlightblue}]{hyperref} 

\usepackage[capitalize]{cleveref}
\crefname{section}{Sec.}{Secs.}
\Crefname{section}{Section}{Sections}
\Crefname{table}{Table}{Tables}
\crefname{table}{Tab.}{Tabs.}

\DeclareMathOperator{\tr}{tr}

\newcommand{\fig}[1]{Fig.~\ref{#1}}

\newcommand{\tbl}[1]{Tab.~\ref{#1}}

\newcommand{\mypar}[1]{\vspace{-3mm}\paragraph{\normalfont\bf #1}\ \ }
\newcommand{\mysection}[1]{\vspace{-1.0mm}\section{#1} }
\newcommand{\mysubsection}[1]{\vspace{-1.0mm}\subsection{#1} }
\def\upvspacefig{\vspace{-1mm}}
\def\vspacefig{\vspace{-3mm}}

\definecolor{lightyellow}{RGB}{255,255,170}
\newcommand{\bestcell}{\bf}

\newcommand{\supparxiv}[2]{#2}

\usepackage{etoolbox}
\makeatletter
\AfterEndEnvironment{algorithm}{\let\@algcomment\relax}
\AtEndEnvironment{algorithm}{\kern2pt\hrule\relax\vskip3pt\@algcomment}
\let\@algcomment\relax
\newcommand\algcomment[1]{\def\@algcomment{\footnotesize#1}}
\renewcommand\fs@ruled{\def\@fs@cfont{\bfseries}\let\@fs@capt\floatc@ruled
  \def\@fs@pre{\hrule height.8pt depth0pt \kern2pt}%
  \def\@fs@post{}%
  \def\@fs@mid{\kern2pt\hrule\kern2pt}%
  \let\@fs@iftopcapt\iftrue}
\makeatother

\makeatletter
\def\adl@drawiv#1#2#3{%
        \hskip.5\tabcolsep
        \xleaders#3{#2.5\@tempdimb #1{1}#2.5\@tempdimb}%
                #2\z@ plus1fil minus1fil\relax
        \hskip.5\tabcolsep}
\newcommand{\cdashlinelr}[1]{%
  \noalign{\vskip\aboverulesep
           \global\let\@dashdrawstore\adl@draw
           \global\let\adl@draw\adl@drawiv}
  \cdashline{#1}
  \noalign{\global\let\adl@draw\@dashdrawstore
           \vskip\belowrulesep}}
\makeatother

\let\subparagraph\paragraph
\usepackage{titlesec}
\titlespacing\section{0pt}{10pt plus 1pt minus 1pt}{5pt plus 1pt minus 1pt}
\titlespacing\subsection{0pt}{8pt plus 1pt minus 1pt}{4pt plus 1pt minus 1pt}
\begin{document}
\pagestyle{headings}
\mainmatter
\def\ECCVSubNumber{4119}  %

\title{Sound Localization by \\ Self-Supervised Time Delay Estimation}

\def\eg{\emph{e.g.}} \def\Eg{\emph{E.g.}}
\def\ie{\emph{i.e.}} \def\Ie{\emph{I.e.}}
\def\cf{\emph{c.f.}} \def\Cf{\emph{C.f.}}
\def\etc{\emph{etc}} \def\vs{\emph{vs.}}
\def\wrt{w.r.t.} \def\dof{d.o.f.}
\def\etal{\emph{et al.}}

\newcommand{\andrew}[1]{\textcolor{blue}{Andrew: #1}}
\newcommand{\ziyang}[1]{\textcolor{red}{Ziyang: #1}}
\newcommand{\david}[1]{\textcolor{green}{David: #1}}

\def\bx{{\mathbf x}}
\def\bu{{\mathbf u}}
\def\bv{{\mathbf v}}
\def\bg{{\mathbf g}}
\def\bgh{{\hat {\mathbf g}}}
\def\bxh{{{\hat {\mathbf x }}}}
\def\bh{{\mathbf h}}
\def\bhh{{{\hat {\mathbf h}}}}
\def\bxo{\bx_1}
\def\bxt{\bx_2}

\titlerunning{Sound Localization by Self-Supervised Time Delay Estimation}
\author{Ziyang Chen \and
David F. Fouhey\index{Fouhey, David F.} \and
Andrew Owens}
\authorrunning{Z. Chen et al.}

\institute{
University of Michigan 
}

\maketitle

\begin{figure}
\centering
\vspace{-5mm}
\includegraphics[width=1.0\textwidth]{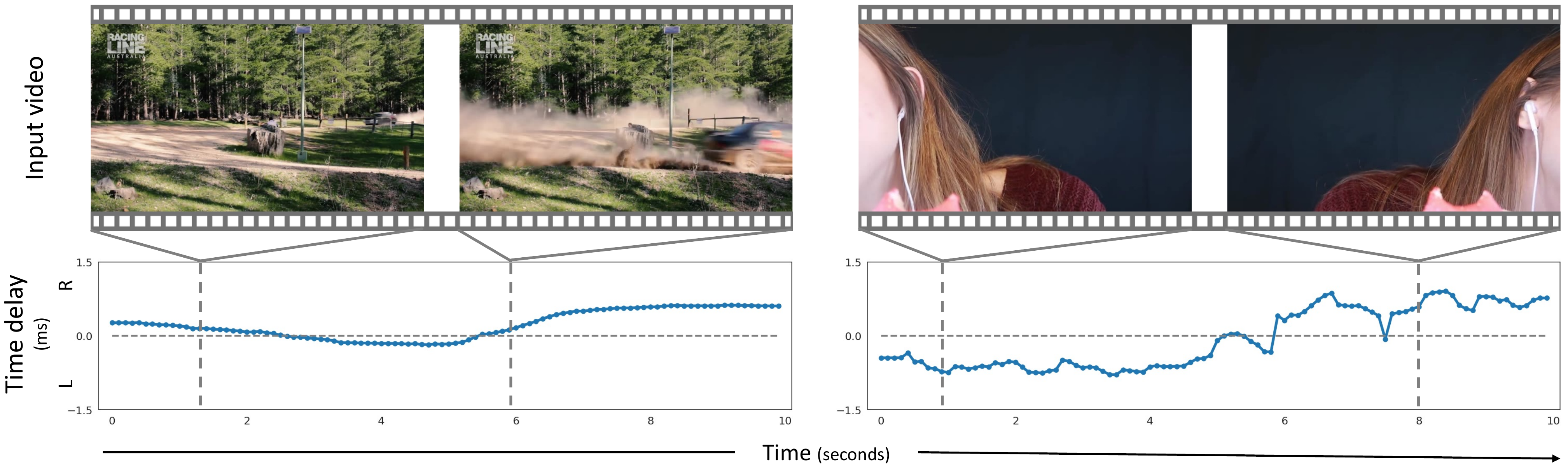}
\raggedright

 \captionof{figure}{%
 Given a stereo audio recording, we estimate a sound's {\em interaural time delay}. Our model learns through self-supervision to find correspondences between the signals in each channel, from which the time delay can be estimated.
 We show time delay predictions for two scenes, along with their corresponding video frames~(not used by the model). In both cases, the sound source changes its position in a scene, resulting in a corresponding change in time delay.} \vspacefig

\label{fig:teaser}
\end{figure}
\vspace{-3mm}

\vspace{-3mm}\begin{abstract}

Sounds reach one microphone in a stereo pair sooner than the other, resulting in an {\em interaural time delay} that conveys their directions.
Estimating a sound's time delay requires finding correspondences between the signals recorded by each microphone. We propose to learn these correspondences through self-supervision, drawing on recent techniques from visual tracking. We adapt the contrastive random walk of Jabri et al.\supparxiv{}{~\cite{jabri2020spacetime}} to learn a cycle-consistent representation from unlabeled stereo sounds, resulting in a model that performs on par with supervised methods on ``in the wild" internet recordings.  We also propose a multimodal contrastive learning model that solves a {\em visually-guided} localization task: estimating the time delay for a particular person in a multi-speaker mixture, given a visual representation of their face. Project site: \url{https://ificl.github.io/stereocrw}.

\end{abstract}

\vspace{-5mm}
\section{Introduction}

Sounds in the world arrive at one of our two ears slightly sooner than the other. This {\em interaural time delay}, which generally lasts only a few hundred microseconds, indicates a sound's direction and thus provides an important cue for multimodal perception. In humans, for example, time delays convey the positions of objects that move out of sight, and are integrated with visual cues when localizing events~\cite{kumpik2019re}. %
Visual information can also guide the sound localization process, allowing us to find a particular event of interest through binaural cues, while ignoring the others. %

While high-quality stereo sound recordings are now abundant, such as in the audio tracks of videos recorded by consumer phones, existing methods often struggle to localize sound sources within them, particularly when they contain correlated noise or multiple sound sources. 
The localization problem has typically been addressed by matching hand-crafted features~\cite{knapp1976generalized,schmidt1986multiple} and, recently, by supervised learning~\cite{houegnigan2017neural,pertila2019time,diaz2020robust}. However, the difficulty in acquiring natural labeled data has limited their effectiveness. Many approaches, consequently, resort to using simulated training data that may not be fully representative of the world.

We propose to address these problems by learning time delay estimation from real, unlabeled recordings. We take inspiration from work in self-supervised visual tracking that learns space-time correspondences from videos, such as through cycle consistency~\cite{wang2019learning,jabri2020spacetime,bian2022learning} and contrastive learning~\cite{wang2021unitrack}. Analogously, our approach is based on learning audio embeddings that can be used to find {\em interaural correspondences}: pairs of sounds from different stereo channels that correspond to the same underlying events.

We introduce a model, inspired by the {\em contrastive random walk} of Jabri et al.~\cite{jabri2020spacetime}, that learns cycle consistent features from unlabeled stereo sound. This model maximizes the return probability of a random walk on a graph whose nodes correspond to the audio samples in each channel. In this graph, edges connect samples between channels, and the walk's transition probabilities are defined by learned embeddings. We show examples of time delay estimates for two real-world videos in Figure~\ref{fig:teaser}.

We also propose a model inspired by {instance discrimination}~\cite{dosovitskiy2015discriminative,wu2018unsupervised,he2019momentum,chen2020simple} that can perform a novel {\em visually-guided} time delay estimation task: localizing a speaker in a multi-speaker audio recording, given only their visual appearance. The resulting model is simple and can accurately localize speakers, without the need for explicitly separating sounds in the mixture. We also use this approach to train audio-based localization models solely from mono audio, which in some domains may be more readily available than stereo sound. This model uses data augmentation to incorporate knowledge about invariances to important sources of variation.

Through experiments on simulated environments with metrically accurate ground truth, and on internet videos with directional judgments annotated by human listeners, we show:
\def\labelitemi{\textbullet }
\begin{itemize}[leftmargin=*,topsep=1pt, noitemsep]
    \item Interaural time delays can be accurately estimated through self-supervised learning, using either unlabeled stereo and mono training data.
    \item Our models provide robustness to distracting sounds within a mixture, and perform well on real-world recordings, obtaining competitive performance with state-of-the-art { supervised} methods. %
    \item Visual signals allow our models to localize specific speakers within mixtures.
\end{itemize}

\mysection{Related Work}
\vspace{3mm}

\mypar{Human binaural localization.} Humans use two main cues for estimating the azimuth of a sound: interaural time differences~(ITD) and interaural intensity differences~(IID), i.e., the difference in the loudness of the sounds entering both ears~\cite{rayleigh1907xii,wang2006computational}. 
In practice, IID is primarily useful for high-frequency sounds that are close to the observer, while ITD is useful for low-frequency sounds and is relatively unaffected by distance~\cite{brungart1998near}. Our work is thus complementary to methods that use IID cues. Humans can accurately estimate azimuth from multi-source mixtures~\cite{yost1996simulated,hawley1999speech}, and integrate vision with binaural cues~\cite{kumpik2019re}, motivating our work on multi-speaker time delay estimation.

\mypar{Time delay estimation.}
Time delay estimation is a classic signal processing problem.
In early work, Carter et al.~\cite{carter1973smoothed}  estimated time delays using generalized cross-correlation with phase transform~(GCC-PHAT), which corresponds to a maximum likelihood estimate under low noise~\cite{knapp1976generalized,brandstein1997practical,zhang2008does}. Other work uses beamforming ~\cite{dibiase2000high} or subspace methods~\cite{schmidt1986multiple}. Comanducci et al.~\cite{comanducci2020time} trained a convolutional network~(CNN) to denoise GCC-PHAT features. Other work trains a multi-layer perceptron to predict a time delay from a raw waveform~\cite{houegnigan2017neural}, trains recurrent networks on hand-crafted features~\cite{pertila2019time,salvati2021time}, and uses 3D CNNs~\cite{diaz2020robust}. Christensen et al.~\cite{christensen2020batvision} used GCC-PHAT and echolocation to estimate depth maps from audio. In concurrent work, Chen et al.~\cite{chen2022learning} localized multiple sounds by jointly solving source separation and time delay estimation problems. Time delay estimation also has a wide range of applications and modalities, such as oceanography~\cite{bianco2019machine}, wireless networking~\cite{yang2019ilps,patwari2005locating}, sonar~\cite{carter1981time}, and possibly directional olfaction~\cite{rajan2006rats}. In contrast, we pose time delay estimation as a self-supervised learning problem, and we do not require hand-crafted features or labels.

\mypar{Supervised binaural localization.}
Vecchiotti et al.~\cite{vecchiotti2019end} estimated sound direction directly from raw waveforms. Other work uses Short-time Fourier Transform~\cite{yalta2017sound,chakrabarty2017broadband,adavanne2018direction} or beamforming features~\cite{salvati2018exploiting}. Due to the challenge in obtaining labeled data, these methods have largely been trained on synthetic or lab-collected data. In contrast to these approaches, we learn a specific~(but widely useful) cue---the time delay---through self-supervision on natural data. %

\mypar{Audio-visual binaural learning.} Yang et al.~\cite{yang2020telling} distinguished between audio-visual examples in which the stereo channels have~(or have not been) swapped, resulting in a representation that can be finetuned to solve localization tasks. In contrast, our model can optionally be trained and deployed solely with audio, and produces an output---the time delay---that is directly correlated with sound direction, without the need for finetuning. Gan et al.~\cite{gan2019self} used a car detector to provide pseudo ground truth for a sound-based localization method. Since the training data comes from a supervised car detector, the model relies on labeled training data, whereas ours is self-supervised. Later work~\cite{dai2021binaural,valverde2021there} extends this approach by distilling supervision from multiple visual classifiers and modalities. Other work generates stereo sound from mono audio using images~\cite{gao20192,garg2021geometry,xu2021visually}, largely by adjusting the relative volume of the channels to simulate IID cues.

\mypar{Audio-visual sound localization and separation.}
A variety of methods have been proposed for using vision to localize and separate sounds. Classic work searches for cross-modal similarity in statistical models~\cite{hershey1999audio,fisher2000learning,kidron2005pixels}. Later work uses contrastive learning to find image regions that are highly correlated with sound~\cite{senocak2018learning,arandjelovic2017objects,owens2018ambient,owens2018audio,zhao2018sound}, and separates sounds from synthetic mixtures~\cite{ephrat2018looking,afouras2018conversation,owens2018audio,gao2018learning,gabbay2018seeing}. Recent work has applied the contrastive random walk to localize multiple sounds within images~\cite{hu2022mix}. This method learns correspondences between image patches and (mono) audio, whereas our model learns correspondences between the signals in each stereo channel.

\mypar{Audio self-supervision.}
A variety of methods have been proposed for learning audio representations through self-supervision, typically for semantic recognition tasks, such as music or speech understanding. These include contrastive learning~\cite{van2018representation,schneider2019wav2vec,wang2021multi,jiang2020speech,gong2021ssast,wang2021towards}, autoencoding~\cite{eloff2019unsupervised}, multi-task learning with pretext tasks~\cite{pascual2019learning}, and generative autoregressive models~\cite{chung2019unsupervised}. In contrast, we learn a representation for learning interaural correspondence in binaural audio.

\mypar{Learning visual correspondences.} 
We take inspiration from methods that learn space-time correspondences from video. These include methods that colorize grayscale video~\cite{vondrick2018tracking}, cycle-consistent feature representations~\cite{wang2019learning,jabri2020spacetime,hadji2021representation} and slow features~\cite{wiskott2002slow,gordon2020watching}. Other work~\cite{wang2021unitrack} has shown that features learned through instance discrimination~\cite{chen2020simple,wu2018unsupervised,dosovitskiy2014discriminative,he2019momentum} are effective for tracking. However, these methods have not been applied to learning stereo audio correspondences. In our models, by contrast, vision is used to aid the audio matching process. We adapt several of these methods~\cite{jabri2020spacetime,gordon2020watching,chen2020simple,bian2022learning} to learn correspondences between temporal samples of audio for binaural matching.

\mysection{Method}

The goal of the time delay estimation problem is to determine how much sooner a sound reaches one microphone than another\footnote{This quantity is also known as the {\em time difference of arrival}~(TDOA) or alternatively as the {\em interaural time difference} or {\em delay}~(ITD).}.
Given the two channels of a stereo recording, $\bxo, \bxt \in \mathbb{R}^n$, represented as waveforms, and a function $h: \mathbb{R}^n \mapsto \mathbb{R}^{n \times d}$ that computes features for each temporal sample, a common solution is to choose a time delay $\tau$ that maximizes the generalized cross-correlation~\cite{knapp1976generalized}:
\vspace{-0.5em}
\begin{equation}
  R_{\bxo, \bxt}(\tau) = \mathbb{E}_t \left[\bh_1(t) \cdot \bh_2(t-\tau)\right],
  \label{eq:cross}
 \vspace{-0.4em}
\end{equation}
where $\bh_i = h(\bx_i)$ are the features for $\bx_i$, and $\bh_i(t)$ is the $d$-dimensional feature embedding for time $t$. 

Traditionally, the audio features, $h$, are defined using hand-crafted features. For example, the widely-used Generalized Cross Correlation with Phase Transform~(GCC-PHAT)~\cite{knapp1976generalized} whitens the audio by dividing by the magnitude of the cross-power spectral density. This approach provides the maximum likelihood solution under certain ideal, low-noise conditions~\cite{knapp1976generalized,brandstein1997practical,zhang2008does}.

We propose, instead, to learn $h$ through self-supervision from unlabeled data. These features ought to capture {interaural correspondences}: observations in both waveforms that were generated by the same underlying events should be close in embedding space. We consider models that can be trained solely from unlabeled stereo or mono sound~(Sec.~\ref{sec:interaural}), or that learn to perform visually-guided estimation from audio-visual data~(Sec.~\ref{sec:visual}).

\mysubsection{Learning interaural correspondence} \label{sec:interaural}
\vspace{-1mm}
We propose models that learn interaural correspondence from unlabeled data.

\begin{figure*}[t]
    \centering
    \upvspacefig
    \includegraphics[width=\linewidth]{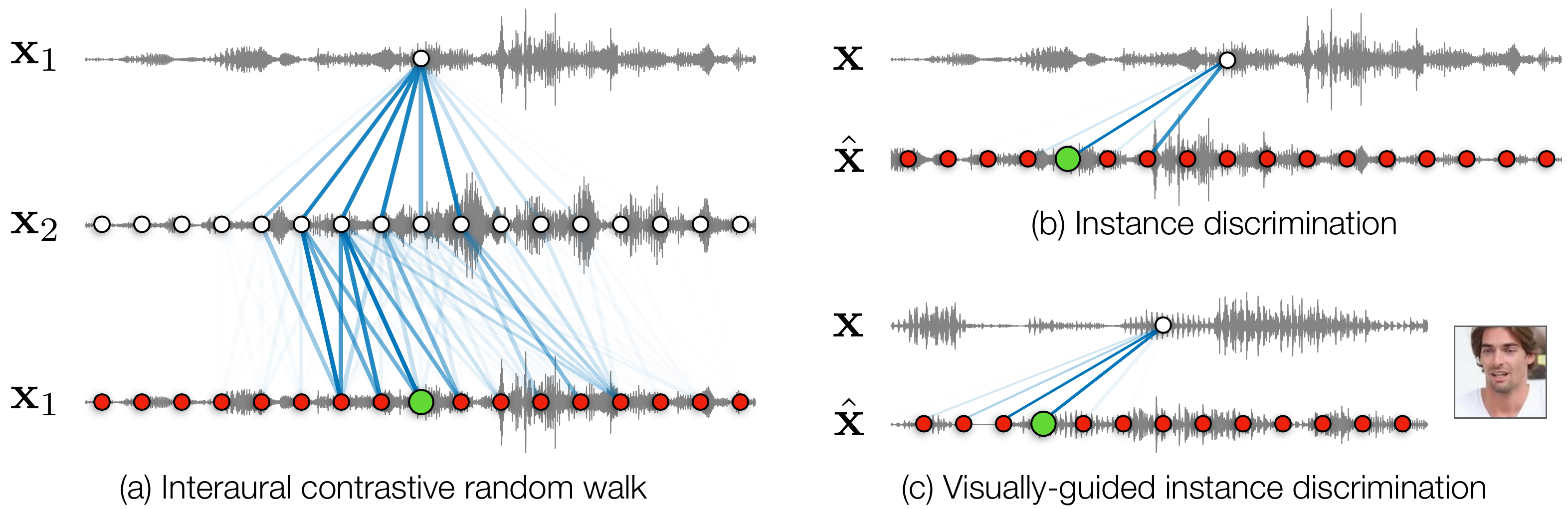}
    \caption{{\bf Learning interaural correspondence}. We consider several self-supervised models. (a) A random walk moves from one stereo channel to another, then back, with transition probabilities determined by our learned embeddings. We learn to maximize the probability that it returns to the node where it started (marked in green). (b) We apply data augmentation to mono audio, and learn embeddings that maximize the similarity of corresponding timesteps. (c) We learn to match audio for a single speaker from a multi-speaker mixture, given a visual input. } 
    \label{fig:method}
    \vspacefig
\end{figure*}

\mypar{Contrastive random walks.}
Our embeddings should provide {\em cycle consistent} matches: the process of matching features from $\bxo$ to those in $\bxt$ should yield the same correspondences as matching in the opposite direction, from $\bxt$ to $\bxo$. We use this idea to learn a representation from unlabeled stereo sounds.

We adapt the contrastive random walk model of Jabri et al.~\cite{jabri2020spacetime} to binaural audio~(\fig{fig:method}a). We create a graph that contains nodes for each of the temporal sample $\bx_i(t)$ from both channels, with edges connecting the nodes that come from different channels.\footnote{Following visual tracking work~\cite{jabri2020spacetime}, one could potentially extend this approach to  microphone arrays with 3 or more channels by performing the walk over all channels. } We then perform a  random walk that transitions from nodes in $\bxo$ to those in $\bxt$, then back to $\bxo$, with transition probabilities that are defined by dot products between embedding vectors:
\vspace{-0.4em}
\begin{equation}
  A_{ij}(s, t) = \frac{\exp(\bh_i(s) \cdot \bh_j(t)/c)}{\sum_{k=1}^n \exp(\bh_i(s) \cdot \bh_j(k)/c)},
  \label{eq:affinity}
\vspace{-0.3em}
\end{equation}
where $A_{ij}(s, t)$ is the probability of transitioning from sample $s$
in $\bx_i$ to sample $t$ in $\bx_j$, and a temperature
constant $c$. The features $\bh_i = h(\bx_i; \theta)$ are parameterized with network weights $\theta$ and are represented using a CNN~(Sec.~\ref{sec:implementation}).
We maximize the log return probability of a walk that moves between nodes in the two channels:
\vspace{-0.4em}
\begin{equation}
  \mathcal{L}_{\mathtt{crw}} = -\frac{1}{n}\tr(\log(A_{12} A_{21})),
  \label{eq:crwloss} 
  \vspace{-0.3em}
\end{equation}
where the $\log$ is computed element-wise. 
We also found it helpful to incorporate knowledge about invariances to important sources of variation, such as to noise. To do this, we also apply data augmentation to two audio channels during the walk similar to Hu et al.~\cite{hu2022mix}~(\supparxiv{see supp. for details}{see Sec.~\ref{appendix:implement} for details}). 

\mypar{Slow features.} We also train a variation of the model that learns to associate embeddings that temporally co-occur, taking inspiration from methods that learn slow features in video~\cite{wiskott2002slow,gordon2020watching} and audio-visual synchronization~\cite{chung2016out,owens2018audio,korbar2018cooperative}. These pairs of embeddings are more likely~(than misaligned timestamps) to correspond to the same events. We minimize:
\vspace{-0.7em}
\begin{equation}
  \mathcal{L}_{\mathtt{zero}} = -\frac{1}{n}\tr(\log(A_{12})),\vspace{-0.7em}
\end{equation}
where $A_{12}$ is defined as in Eq.~\ref{eq:affinity}.

\mypar{Instance discrimination.} We also consider models that can be trained solely with mono audio using {\em instance discrimination}~\cite{wu2018unsupervised}. In lieu of a second audio channel, we create synthetic views of mono audio, using data augmentation that encourages invariances that are likely to be useful for interaural matching. We minimize:
\vspace{-0.4em}
\begin{equation}
  \mathcal{L}_{\mathtt{dis}} = -\log  \frac{\exp(\bh(t) \cdot {\bhh}(t)/c)}{\sum_{k=1}^n \exp(\bh(t) \cdot \bhh (k)/c)},
  \vspace{-0.3em}
\end{equation}
over all timesteps $t$, where $\bh = h(\bx)$ are the features for a mono audio $\bx$, and ${\bhh} = h(\bxh)$ are features computed from an augmented version of $\bx$.

Unless otherwise specified, we perform two types of augmentation: time shifting and volume adjustment. To model the challenges in time delay estimation, we choose
negative examples exclusively from $\bx$, rather than other examples
in the batch~\cite{chung2016out,owens2018audio,korbar2018cooperative}. We ensure that augmented positive views are always taken from the corresponding timestep, i.e., we undo any time-shifting augmentation when indexing $\bhh(t)$.

\mysubsection{Visually-guided time delay estimation} \label{sec:visual}

We also apply our model to the novel problem of estimating the time delay for a single sound within a mixture using visual information. Given a sound mixture containing multiple simultaneous speakers, we estimate the time delay for one object, given a visual representation of its appearance~(e.g., localizing a speaker using a visual representing their face). The visual input need not co-occur with the audio. For example, the object may be off-screen, or its visual features may have been extracted at an earlier time. 

We adapt the instance discrimination variation of our model, with a training procedure that resembles the ``mix-and-separate''~\cite{zhao2018sound} paradigm used in audio-visual source separation~\cite{ephrat2018looking,afouras2018conversation,owens2018audio,gao2018learning,gabbay2018seeing}. We create a mixture from two sounds, each with its own delay, and ask the model to estimate the delay from only the desired source. Given two audio tracks $\bu$ and $\bv$, we create a synthetic binaural sound mixture $\bx_1 = \bu + \bv$ and $\bx_2 = \mathrm{shift}(\bu, \tau_u) + \mathrm{shift}(\bv, \tau_v)$ for randomly sampled values $\tau_u$ and $\tau_v$, where $\mathrm{shift}(\bx, \tau)$ shifts $\bx$ by $\tau$. The model is also provided with $I_u$, an image depicting $\bu$. We learn audio-visual features by minimizing:
\vspace{-0.4em}
\begin{equation}
  \mathcal{L}_{\mathtt{av}} = -\log \frac{\exp(\bg_1(t) \cdot {\bg_2}(t - \tau_u)/c)}{\sum_{k=1}^n \exp(\bg_1(t) \cdot \bg_2 (k)/c)},
  \vspace{-0.3em}
\end{equation}
over all timesteps $t$, where $\bg_i = g(\bx_i, I_u)$ are the learned audio-visual features for channel $\bx_i$. Here, $g$ obtains its embedding by fusing audio from one channel with the input image. As in the instance discrimination model, we apply augmentation to $\bg_2$. Note that this task cannot be solved without $I_u$: from audio alone, the model would be unable to determine whether the true delay is $\tau_u$ or $\tau_v$.

\mysubsection{Learning a time delay estimation model} \label{sec:implementation}

We now describe how these self-supervised learning models can be trained, and how they can be used to estimate time delays.

\mypar{Network architectures.} We implement the audio embedding $h$ using a CNN that operates on spectrograms~\cite{hershey2017cnn}. To compute the embedding for sample $s$, we extract a waveform of length $T$ centered on $s$. We create a spectrogram representation of size $128 \times 128 \times 2$ using a Short-time Fourier transform~(STFT). We keep both magnitude and phase and provide them as input to a ResNet~\cite{he2016}, which extracts a $d = 128$ dimensional, $\ell_2$-normalized embedding. %

For our audio-visual model, we represent the visual information using~(pretrained) FaceNet~\cite{schroff2015facenet}. This allows our model to estimate attributes of speakers from face crops, similar to the work in source separation that uses face embeddings~\cite{ephrat2018looking}. We fuse the audio and visual features after the second convolution block of the audio subnetwork by concatenating the 128-dimensional visual features at each time-frequency position. 

\mypar{Datasets.}
We train our audio models on datasets of stereo sound: {\bf FAIR-Play}~\cite{gao20192}, which has 1,871 videos~(5.2 hours) of lab-collected music performances from a small number of rooms %
and {\bf Free-Music-Archive}~(FMA)~\cite{defferrard2016fma}, a dataset of 101K~(841 hours) music recordings created by a large number of artists.  %

For the visually-guided model, we train our model on VoxCeleb2~\cite{chung2018voxceleb2} with 500 randomly selected identities. We randomly create training mixtures from mono sound, without the speaker identity labels and without using a simulator.

\mypar{Training.}
We use the AdamW optimizer~\cite{loshchilov2017decoupled,kingma2015adam} with a learning rate $=10^{-4}$, a cosine decay learning rate scheduler, a batch size of 48, a temperature $c = 0.05$ following \cite{wu2018unsupervised}, and early stopping. \supparxiv{Please see supp. for more training details}{Please see Sec.~\ref{appendix:implement} for more training details}.

\mypar{Self-supervised learning formulation.}
For all models, we extract our examples from a 1220-sample waveform, sampled at 16 Khz. We obtain our embeddings using a sliding window of size 0.064s~(1024 samples), with a step of 4 samples, yielding 49 audio clips. 
We apply random stereo channel swapping and channel-wise waveform rescaling  for augmentation in all models. In some experiments, we also add random noise, add reverberation, and mix in other sounds as additional augmentation. At test time, we obtain a denser audio graph by using the step of 1 sample for the sliding window. Our model can use input sounds with a variety of durations without retraining, due to the fully-convolutional network architecture~\cite{long2015}. 
For the audio-visual task, we use a window length of 0.96s or 2.55s to obtain more temporal context, since this problem involves jointly solving a separation task.

\mypar{Estimating delays from features.} \textls[0]{After learning our representation $h$, we can use it to estimate the time delay, such as by maximizing $R_{\bxo, \bxt}$~(Eq.~\ref{eq:cross}). We have found that this procedure can affect the quality of the prediction for both learned and hand-crafted methods~(e.g., due to outliers), so we evaluate a number of different variations in our experiments. In our approach, each embedding votes on a value for $\tau$. We then choose a single time delay for the audio from these votes, either by taking the mean or by using a RANSAC-like~\cite{fischler1981random} mode estimation method. %
In the latter, we first select the delay with the most votes, then average the inliers~(those within a small threshold of the chosen value). This vote can be performed by the nearest neighbor search, or by treating the learned similarities as probabilities~(Eq.~\ref{eq:affinity}) and taking the expectation, \ie, $\frac{1}{n}\sum_{s, \tau} \tau A_{12}(s, \tau)$ ~\cite{bian2022learning}. }  %

\mysection{Experiments}

We evaluate our methods using both simulated audio, where time delays can be measured exactly, and real-world binaural audio from unknown microphone geometry, where quantized sound direction categories are labeled by humans. %

\mysubsection{Evaluation with Simulated Sounds} \label{sec:simulate}
Before considering real-world audio, we evaluate each model's performance on time delay estimation task using simulated environments, following \cite{salvati2021time}. While the resulting sounds are considerably simpler than real-world recordings, they allow us to obtain metrically accurate ground-truth time delays, and to systematically vary different experimental conditions, such as the amount of background noise.

\mypar{Simulation.} 
Following previous work~\cite{salvati2021time}, we simulate stereo sounds using Pyroomacoustics~\cite{scheibler2018pyroomacoustics}.  We create three simulated environments with rooms of different sizes and microphone positions. For our sound sources, we take speech sounds from TIMIT~\cite{garofolo1993timit}~(recorded in anechoic conditions) and place them at random angles sampled uniformly from (-90$^\circ$, 90$^\circ$) and distances (0.5m, 3.0m) with respect to the microphone. We add independent Gaussian noise to create conditions with different signal-noise ratio~(SNR) levels, and consider a variety of reverberation times~($\text{RT}_{60}$). The ground-truth time delay can straightforwardly be calculated from the sound source and the microphone pose. This simulated test set, which we call \textbf{TDE-Simulation}, contains approximately 6K audio samples total~(\supparxiv{please see the supplement for more details}{please see Sec.~\ref{appendix:sim} for more details}). 

\mypar{Models.}  We evaluated our audio-based learning methods: 1) {\bf StereoCRW}, contrastive random walks trained on stereo sounds, 2) {\bf ZeroNCE}, slow features trained on stereo sounds~(named after VINCE~\cite{gordon2020watching}), and 3) {\bf MonoCLR}, and instance discrimination trained on mono sounds~(named after SimCLR~\cite{chen2020simple}).

We compared our methods with the widely-used {\bf GCC-PHAT}~\cite{knapp1976generalized}, a hand-crafted audio feature. We also compared with the recent {\em supervised} method {\bf Salvati et al.}~\cite{salvati2021time}, which trains a CNN on parameterized GCC-PHAT features to regress time delay. We trained this model on simulated stereo sounds, based on audio clips from VoxCeleb2~\cite{chung2018voxceleb2} to obtain human speech signals. To improve this baseline's performance, we make a modification: in addition to the noise and reverberation augmentations from~\cite{salvati2021time}, we train with synthetic sound mixtures, in which a background sound is added to the input waveform. We regard this supervised method as an approximate upper bound for the simulation-based experiments. We provide all methods with the same duration audio as input, and evaluate different post-processing methods.

\begin{wraptable}{r}{0.52\textwidth}
\vspace{-2.2em}
\caption{{\bf Delay estimation on TDE-Simulation data.} \textls[-9]{We use SNR=10 and $\text{RT}_{60}$=0.5s. FAIR is FAIR-Play~\cite{gao20192}, FMA is FreeMusic-Archive~\cite{defferrard2016fma}. \emph{Vox-Sim} is the simulator~\cite{scheibler2018pyroomacoustics} with VoxCeleb2~\cite{chung2018voxceleb2} clips and \emph{FMA-Sim} is the simulator with FreeMusic-Archive clips. Errors in ms. {\em Sup} refers to supervision, and {\em Aug} refers to augmentation.}}

 \label{tab:estimate_itd}
 \centering
 \resizebox{1.0\linewidth}{!}{
 \begin{tabular}{ll@{\hskip10pt}lccccc}
 \toprule
 Model & Variation & Data & Sup & Aug & MAE & RMSE \\
 \midrule
 \multirow{4}{*}{{\small Salvati et al.}~\cite{salvati2021time}} 
 & {\small Mean} & Vox-Sim & \checkmark & & {\bestcell 0.126 } & {\bestcell 0.254} \\
 & {\small Mean} & Vox-Sim & \checkmark & \checkmark & { 0.169} & { 0.294} \\
 
 & {\small Mean} & FMA-Sim & \checkmark & & 0.135 & 0.256 \\
 & {\small Mean} & FMA-Sim & \checkmark & \checkmark & { 0.146} & {0.267} \\
 \midrule
 \multirow{2}{*}{{\small GCC-PHAT}~\cite{knapp1976generalized}} & {\small Mode} & \multicolumn{1}{c}{--} & & & 0.179 & 0.396 \\
 & {\small Mean} & \multicolumn{1}{c}{--} & & & {\bestcell 0.160} & {\bestcell 0.318} \\

 \midrule
 \multirow{13}{*}{Ours}
 & {\small Random} & \multicolumn{1}{c}{--} & & & 0.448 & 0.505 \\
 \cdashlinelr{2-7}
 & {\small MonoCLR} & FAIR & & & 0.395 & 0.566 \\ 
 & {\small MonoCLR} & FAIR & & \checkmark & 0.202 & 0.340 \\ 
 & {\small ZeroNCE} & FAIR & & & 0.241 & 0.362 \\ 
 & {\small ZeroNCE} & FAIR & & \checkmark & 0.196 & 0.366 \\ 
 & {\small StereoCRW} & FAIR & & & 0.241 & 0.364 \\ 
 & {\small StereoCRW} & FAIR & & \checkmark & 0.174 & 0.322 \\ 
 \cdashlinelr{2-7}
 & {\small MonoCLR} & FMA & & & 0.430 & 0.648 \\ 
 & {\small MonoCLR} & FMA & & \checkmark & 0.187 & 0.335 \\ 
 & {\small ZeroNCE} & FMA & & & 0.227 & 0.347 \\ 
 & {\small ZeroNCE} & FMA & & \checkmark & 0.174 & 0.319 \\ 
 & {\small StereoCRW} & FMA & & & 0.434 & 0.654 \\ 
 & {\small StereoCRW} & FMA & & \checkmark & {\bestcell 0.133} & {\bestcell 0.259} \\ 
 \bottomrule

 \end{tabular}
}

\vspace{-2.2em}
\end{wraptable} %
\mypar{Evaluation with moderate noise.} We first evaluate our models on TDE-Simulation with $\text{SNR}=10$ and $\text{RT}_{60}=0.5\text{s}$, a condition with moderate amounts of noise and reverberation. This simulated setup is also well-suited for analyzing 
traditional techniques~\cite{knapp1976generalized,zhang2008does}, which are designed to deal with unstructured, independent noise and reverberation.
We evaluate three audio-based variations of our model, with and without augmentation, and on the different unlabeled training sets. For the instance discrimination model, we always include time-shifting, since the model cannot be trained without some form of augmentation.
To measure prediction accuracy, we use mean absolute error~(MAE) and root mean square error~(RMSE) in milliseconds~(ms). For all the methods, we provide 1024~(0.064s) audio samples~(at 16Khz) as input and perform 128 time delay prediction votes~(Sec.~\ref{sec:implementation}). For GCC-PHAT and Salvati et al., we combine the votes into a single prediction by computing the mode or mean. For our method, we use the mode. We provide an ablation study about post-process, input duration and data distribution gap in \supparxiv{the supplement}{Sec.~\ref{appendix:ablation}}.

\begin{table}[b]
\vspace{-3.5em}
    \centering
    \begin{minipage}[c][1\width]{0.49\textwidth}
        \captionsetup{type=figure}
        
\centering

    \includegraphics[width=1.0\columnwidth]{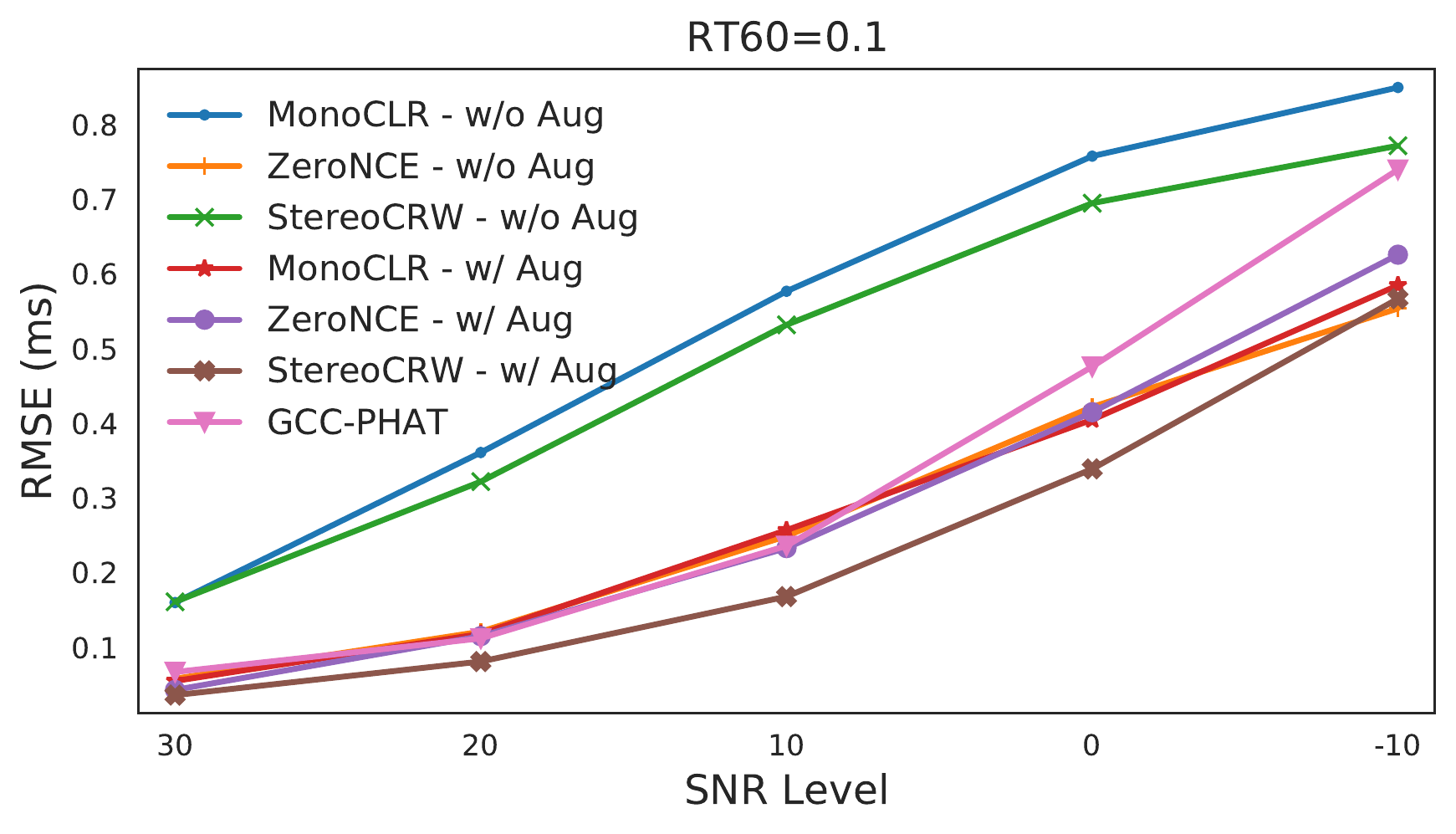}
    \caption{ Robustness to random noise.}
    \label{fig:noise_robust}

    \end{minipage}
    \hfill
    \begin{minipage}[c][1\width]{0.49\textwidth}
    \captionsetup{type=figure}
            \centering
    \includegraphics[width=1.0\columnwidth]{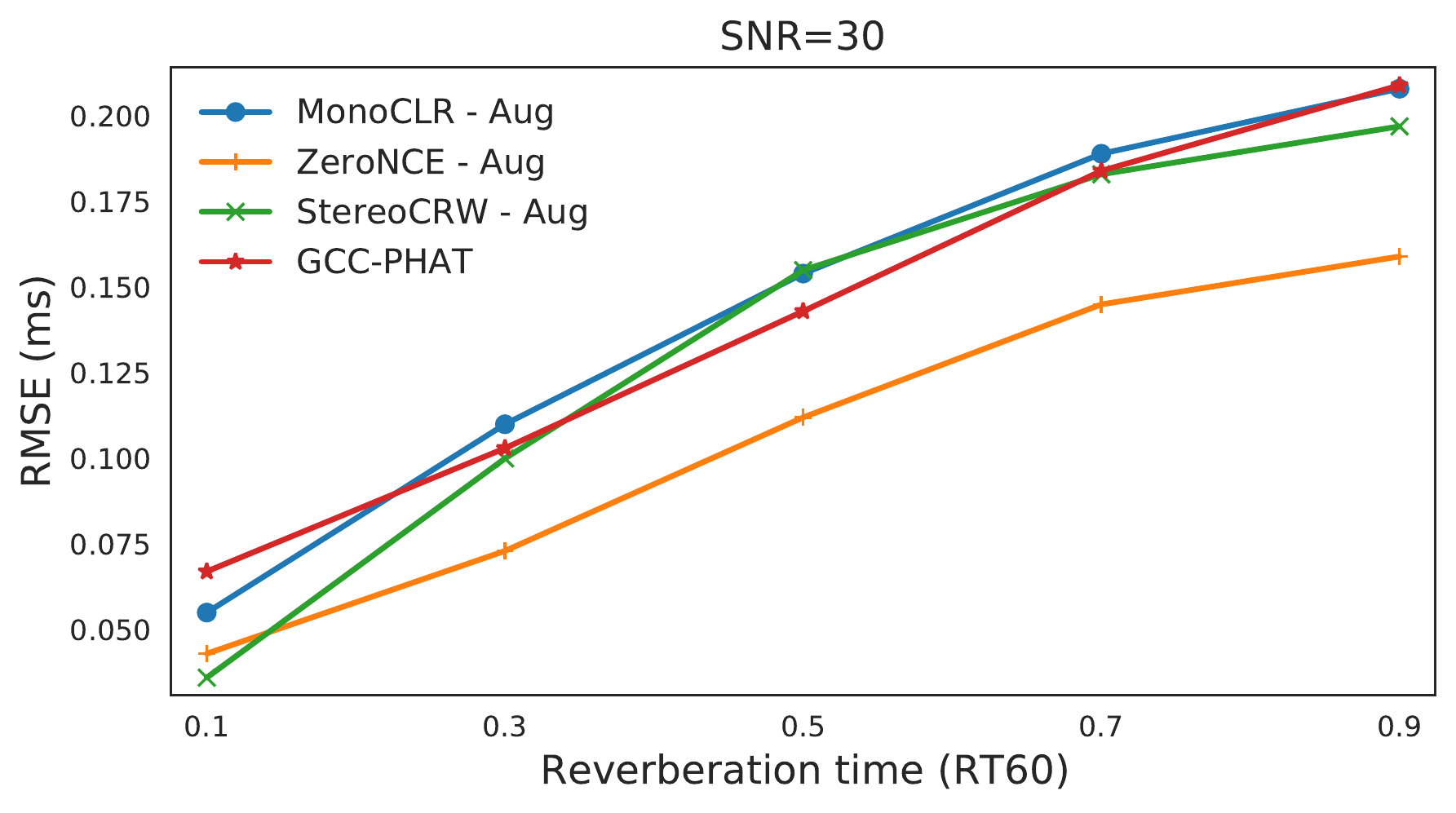}
    \caption{Robustness to reverberation.}
    \label{fig:rt_robust}

    \end{minipage}
\vspace{-5em}
\end{table}

As shown in \tbl{tab:estimate_itd}, the StereoCRW model substantially outperforms GCC-PHAT when it is trained on a large stereo dataset, FreeMusic-Archive, obtaining performance comparable with supervised models trained on synthetic data. 
While ZeroNCE is trained with real stereo sound, its loss implicitly assumes the true time delay is zero, which is violated in real scenes. The cycle consistency loss in StereoCRW does not make this assumption, allowing it to learn from more complex data and learn better representations. The ZeroNCE model outperforms MonoCLR, suggesting that stereo sounds are useful training signals. Augmentations are important for all the models. 
We also note that the data distribution between training and test cases is quite different (\ie, training with music signals and testing on the human speech), suggesting that our approach is capable of generalization.

 \begin{table}[t]
 \vspace{-2mm}
\centering
 \caption{{\bf Sound mixtures on TDE-Simulation data.} We evaluate time delay estimation with sound mixtures under simulator settings $\text{SNR}=30$ and $\text{RT}_{60}=0.1s$. We report RMSE~(ms). }
  \label{tab:mixitd}
 \resizebox{0.9\linewidth}{!}
 { 
  \begin{tabular}{l@{\hskip10pt}l@{\hskip10pt}c@{\hskip10pt}c@{\hskip10pt}c@{\hskip10pt}c@{\hskip10pt}c@{\hskip10pt}c@{\hskip10pt}c}
    \toprule
     &  &  & & \multicolumn{5}{c}{Intensity level of distracting sound}\\
    \midrule
     Model   & Variation & Aug & Dataset &  0.1 & 0.3 & 0.5 & 0.7 & 0.9  \\ 
    \midrule

    \multirow{4}{*}{Salvati et al.~\cite{salvati2021time}} &  Mean  &   & Vox-Sim   & 0.030 & 0.068 & 0.171 & 0.298 & 0.415 \\

    &  Mean  & \checkmark & Vox-Sim & {\bestcell 0.027} & {\bestcell 0.055 } & {\bestcell 0.092} & {\bestcell 0.200} & {\bestcell 0.370} \\
    &  Mean  &   & FMA-Sim   & 0.047  & 0.091  & 0.196 &  0.326 &  0.423 \\
    
    &  Mean  & \checkmark & FMA-Sim   & 0.051  & 0.073  & 0.102 & 0.217  &  {\bestcell 0.370}\\
    \cmidrule{1-9}
    GCC-PHAT~\cite{knapp1976generalized} & Mean  &  &  -- & {\bf 0.024}    & 0.177  &  0.304 & 0.395 &  0.459 \\
    \cdashlinelr{1-9}
    \multirow{6}{*}{Ours} & 
    MonoCLR  &   &  FMA & 0.078   & 0.204 &  0.302  & 0.389  & 0.452 \\
    & MonoCLR  & \checkmark  &  FMA  & 0.055 &   0.103 & 0.185 & 0.306 & 0.427 \\
    & ZeroNCE  &  &  FMA    &   0.065 &  0.129  & 0.212 &  0.294 & {\bf 0.369}\\
    & ZeroNCE  & \checkmark &  FMA & 0.091 & 0.164 & 0.249 & 0.332 & 0.405 \\
    & StereoCRW  &   &  FMA &   0.052 &  0.138  & 0.254  & 0.358 & 0.438 \\
    & StereoCRW  & \checkmark &  FMA  & 0.041  &  {\bf 0.079}  & {\bf 0.144} &  {\bf 0.273} & 0.417\\
    \bottomrule
  \end{tabular}
  }
  \vspace{-3mm}
  \end{table}

\mypar{Robustness to noise and reverberation.}
Following \cite{salvati2021time}, we evaluate our model's robustness to noise. We simulate sounds with the fixed reverberation time $\text{RT}_{60}=0.1\text{s}$. We use the models trained on FreeMusic-Archive dataset and evaluate them with different SNR levels. In \fig{fig:noise_robust}, we see that our methods trained with augmentation outperform GCC-PHAT, with a gap that widens as the amount of noise increases. This suggests that augmentation allows us to build in useful invariances that may not be captured by hand-crafted features.

We also evaluated robustness to reverberation. We fixed the SNR level to 30dB and used reverberation conditions $\text{RT}_{60}$ in the range of $[0.1, 0.9]$ via the simulation. As shown in \fig{fig:rt_robust}, our slow feature method outperforms the baseline under each reverberation condition while the other two approaches show similar overall performances as GCC-PHAT. The results suggest our approaches are robust to some amount of reverberation. 

\mypar{Robustness to mixed-in sounds. }
In the previous experiments, the noise in two channels is designed to be random and uncorrelated. However, in real-world audio, a major source of error comes from other sound sources~(\eg, background sounds). These sound sources are also present in the scene at some spatial position, thus generating a correlated error~(and possibly time delay) in both channels.  %

We design experiments to investigate the model's ability to ignore distracting background sounds within a mixture. We create synthetic mixtures  from TDE-Simulation by mixing two sounds with different angles and distances, using $\text{SNR}=30$ and $\text{RT}_{60}=0.1\text{s}$.  We set one sound source to be the ``dominant" signal and rescale the distracting sound to be 10\% -- 90\% loudness of the dominant source~(such that the delay of the louder sound is considered to be the correct answer). We used the models from FMA and evaluate them with 0.5s audio to ensure that there is sufficient signal to identify the dominant sound. 

In \tbl{tab:mixitd}, we see that our proposed approaches significantly outperform GCC-PHAT when the distracting sounds became louder, suggesting that our model has obtained robustness to sound mixtures. For very quiet mixtures~(10\% the volume of the dominant source), GCC-PHAT outperforms our model, which is understandable, given that this domain resembles noise-free audio, which it is well-suited to~\cite{knapp1976generalized}. The supervised model of Salvati et al.~\cite{salvati2021time} performs worse than our method when distracting sounds have a high intensity level~(above 50\%) unless we explicitly add the same mixture augmentation used in our methods.

\begin{figure*}[!t]
    \centering
    \upvspacefig
    \includegraphics[width=\textwidth]{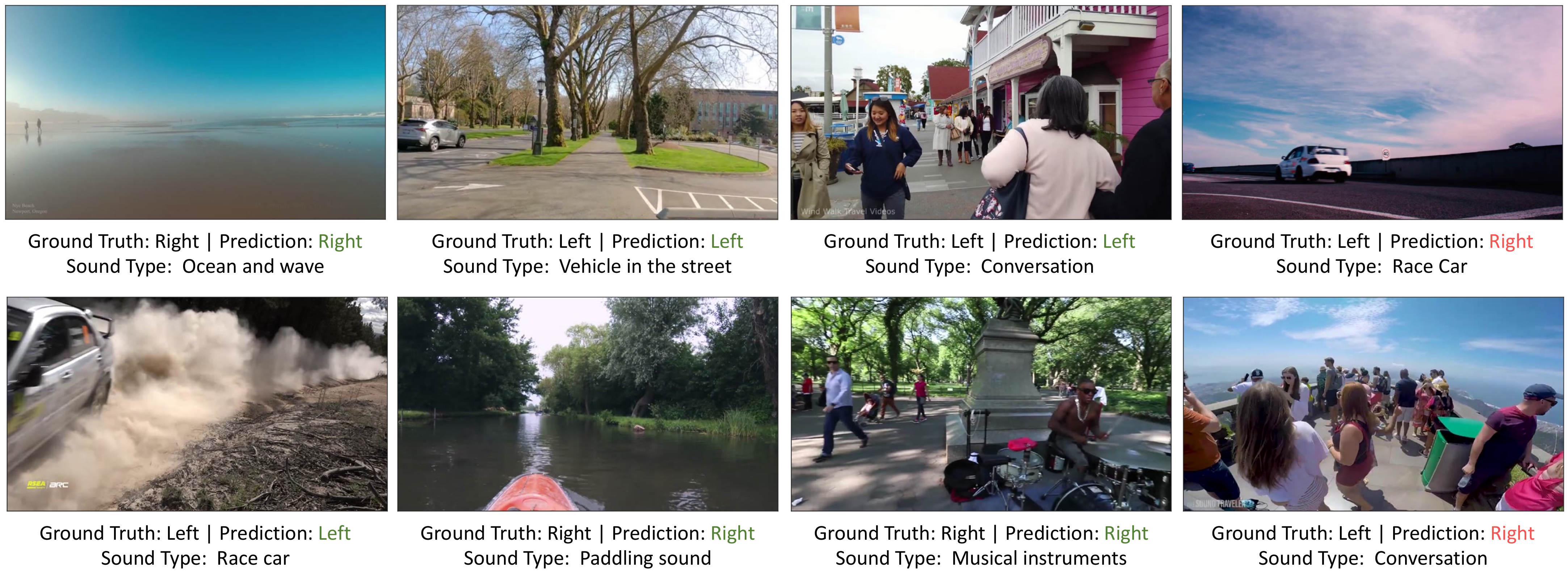}
    \caption{
    {\bf Qualitative results for in-the-wild audio.} We provide an overview of our In-the-wild Binaural dataset along with predictions from our StereoCRW model. \textcolor{green}{Green} denotes the correct predictions and \textcolor{red}{red} means the wrong predictions. We show our failure cases in the last samples on each row. For clarity, we chose examples in which the sound source is evident from the video frame. \supparxiv{Please see supp. for more videos}{More video results can be found on our \href{https://ificl.github.io/stereocrw}{project webpage}}.
    } 
    \label{fig:inthewild}
    \vspacefig
\end{figure*}

\mysubsection{Evaluation with In-the-Wild Audio Recordings}

\textls[0]{Next, we ask how well our time delay estimation methods can localize sound directions in challenging real-world scenes, using audio collected from the internet. }

\mypar{In-the-wild data.}
We collected 30 internet binaural videos, extracted 1K samples from them, and use human judgments to label sound directions. These videos contain a variety of sounds, including engine noise and human speech, which are often far from the viewer. Many also contain multiple sound sources and background noise. \supparxiv{We provide examples of these videos in the supplement}{We provide examples of these videos on our \href{https://ificl.github.io/stereocrw}{project webpage}}.

Since it is difficult for humans to describe sound directions in terms of time delay, we asked listeners to  annotate the direction of the loudest sound. The annotator~(one of the authors) listened to the audio with headphones and labeled 5 directions: {\em left}/{\em right}, {\em center} {\em left}/{\em right}, and {\em center}. %
From these, we created binary {\em left}/{\em right} labels, which can be objectively evaluated by thresholding the delay: we discard the {\em center} label and merge the remaining directional labels~(resulting in 885 examples). We measure the accuracy of the thresholded time delay using these labels. We balance the dataset by swapping stereo channels, such that chance is 50\%.\footnote{In the original version of the paper, we evaluated on the raw (unbalanced) data. Since arXiv v3, we balanced the dataset by swapping stereo channels (Tab.~\ref{tab:inthewild} and \ref{tab:data_ablation}).}
As in the mixture experiments, we provide models with 0.5s audio to ensure that they have sufficient context. We also compared with a  method that uses interaural intensity difference~(IID) cues, by comparing the root mean square~(RMS) of each audio channel to determine which channel is louder to predict its left/right direction~(equivalent to thresholding based on $||\bx||$).

\begin{wraptable}{r}{0.5\textwidth}
\vspace{-2.3em}

\centering
 \caption{{\bf In-the-wild evaluation.} We evaluate our models' ability of localizing sounding objects on \textbf{in-the-wild} test cases.}
  \label{tab:inthewild}
 \resizebox{1.0\linewidth}{!}
 { 
  \begin{tabular}{l@{\hskip10pt}l@{\hskip5pt}c@{\hskip5pt}c@{\hskip5pt}c}
    \toprule
     Model  & Variation & Aug. & Dataset & Acc~(\%)~$\uparrow$   \\ 
    \midrule
    \multirow{4}{*}{Salvati et al.~\cite{salvati2021time}}  &  Mean  &  & Vox-Sim  &  87.7\\
    &  Mean  & \checkmark & Vox-Sim  &  85.8 \\
    &  Mean  &  & FMA-Sim  & {\bestcell 88.0 }\\
    &  Mean  & \checkmark & FMA-Sim  & 87.7 \\
    \cmidrule{1-5}
    Chance & -- & & -- & 50.0 \\
    IID  &  -- &  & --  &  75.4 \\
    GCC-PHAT~\cite{knapp1976generalized}   &  Mean  &  & -- &  77.2	   \\
    \cdashlinelr{1-5}
    \multirow{6}{*}{Ours} 
    & Random  &  --  &   --  & 70.4 \\
    & MonoCLR  &    & FMA  &  83.8  \\
    & MonoCLR  & \checkmark   & FMA   & {\bf 87.4}   \\
    & ZeroNCE  &   & FMA  &   85.4   \\
    & ZeroNCE  & \checkmark & FMA  &    85.6    \\
    & StereoCRW  &   & FMA  &   82.2      \\
    & StereoCRW  & \checkmark & FMA  &  87.2  \\
    \bottomrule
  \end{tabular}
  }   
\vspace{-2.5em}
\end{wraptable} 
\textls[-0]{As shown in \tbl{tab:inthewild}, our proposed approaches all substantially outperform GCC-PHAT. Interestingly, our top-performing model shows comparable results to a state-of-the-art {\em supervised} method, Salvati et al. We also found that augmentation generally improved results, perhaps due to the complexity of the scenes. The IID-based model performed poorly, which may be due to the fact that the sounds were often distant from the camera, and due to the presence of multiple sound sources. We show qualitative results in \fig{fig:inthewild}.}

\begin{figure*}[!t]
    \centering
    \upvspacefig
    \includegraphics[width=\textwidth]{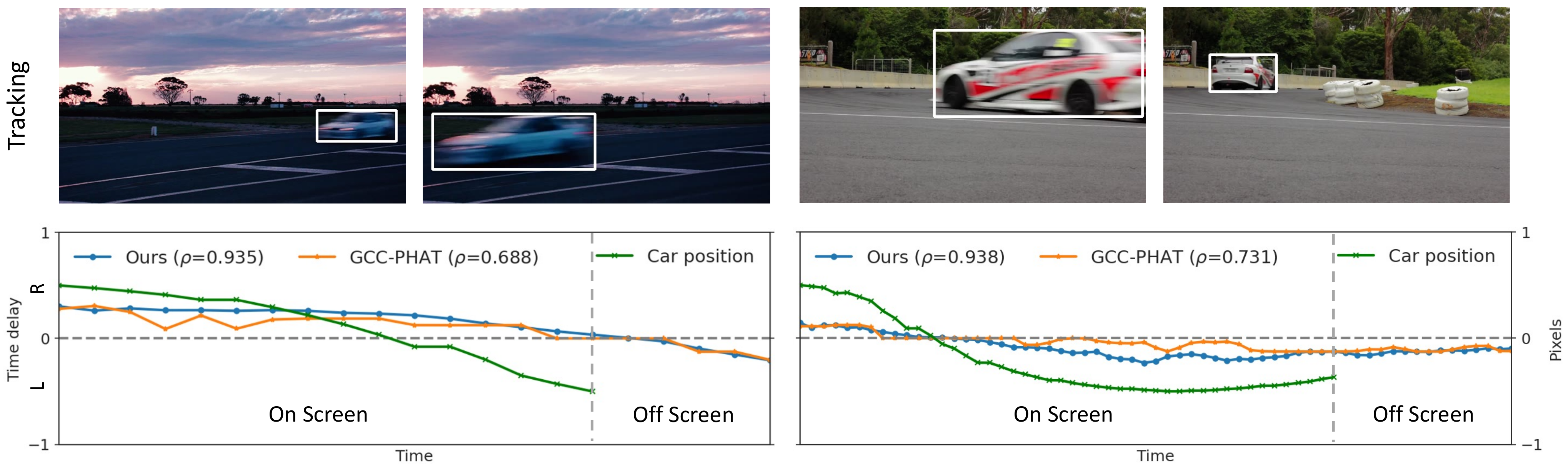}
    \caption{{\bf Motion and time delays.} We show the time delays for both our method and for GCC-PHAT, along with the $x$ coordinate for a tracked vehicle. We also show their correlation. We continue showing the time delay when the car moves off-screen.
    }
    \label{fig:track2itd}
    \vspacefig
\end{figure*}

\mypar{Correlation between visual motion and time delay.}
To help understand how our predicted time delays vary with motion and change over time, we correlated the visual motions in our dataset with the predicted time delays. %
We tracked race cars in the subset of our in-the-wild dataset that contains them using CenterTrack~\cite{zhou2020tracking}, and manually removed erroneous tracks~(obtaining 49 trajectories). We applied our StereoCRW model~(with 1024 samples and 128 votes). In qualitative examples~(\fig{fig:track2itd}), we see that the car's on-screen position is closely correlated with the time delay. Interestingly, our model continues to convey the car's position when it moves off-screen. We computed the Spearman rank correlation coefficient~\cite{spearman1961proof} between the $x$ position of each track~(using the center of the bounding box) and the time delay predictions~(averaged over all video clips), yielding $\rho = 0.57$ for our model and $\rho = 0.48$ for GCC-PHAT. \supparxiv{More video results are shown in the supplement.}{More video results can be found on our \href{https://ificl.github.io/stereocrw}{project webpage}.}

\mypar{Application to phone recordings. }
We also found that our model worked successfully on sounds from ordinary video recordings from recent iPhones. For qualitative results, please see \supparxiv{the supplement}{Sec.~\ref{appendix:iphone_demo}}.

\begin{figure*}[!t]
    \centering
    \upvspacefig
    \includegraphics[width=\textwidth]{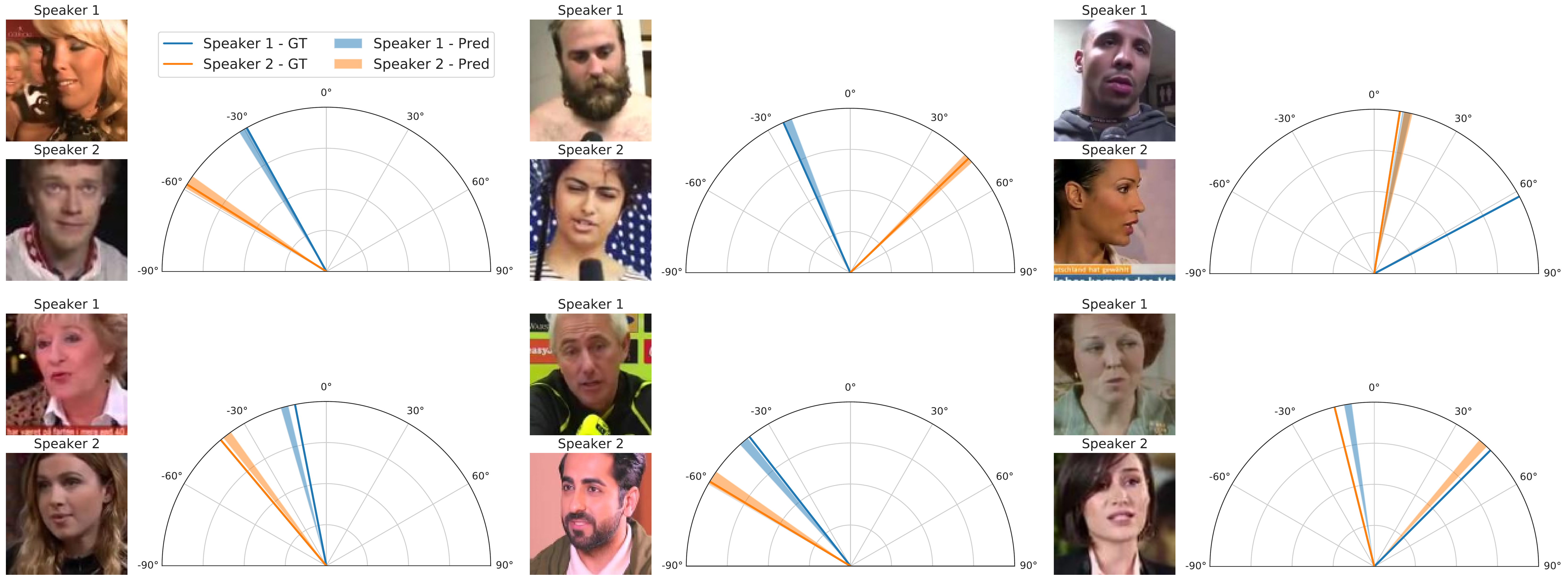}
    \caption{{\bf Qualitative results for visually-guided localization.} We estimate the azimuth of speakers by converting time delay predictions to angles. To convey uncertainty, we show the angular cone corresponding to $\pm 0.25$ samples of error. \textls[-6]{We show failure cases in the last example in each row. \supparxiv{Please see supp. for more video results}{Please see the \href{https://ificl.github.io/stereocrw}{project webpage} for video results}}.
    } 
    \label{fig:speaker_itd}
    \vspacefig
\end{figure*}

\mysubsection{Visually-guided Time Delay Estimation}
Instead of attending to a loud, dominant sound in a mixture~(Sec.~\ref{sec:simulate}), we use vision to specify which, of several, sound sources to localize. The model solves this task by first learning to %
associate a voice with the given visual attribute of the speakers.  Unlike audio spatialization and active speaker detection tasks, which provide temporal and spatial cues from the video, the audio-visual streams in our case are not necessarily temporally aligned (\eg, modeling the challenges of tracking a speaker when they move out of sight).

\mypar{Evaluation dataset.}
We evaluate the audio-visual model in a simulated environment. This allows us to control the positions of the speakers, and to remove other localization cues from the images and audio. We use audio clips from VoxCeleb2~\cite{chung2018voxceleb2} with the  simulation parameters from Sec.~\ref{sec:simulate}. We select 500 speakers from the database~(same as training) and pair them with their corresponding face images. %
Following the work in source separation, we remove loudness cues by normalizing the volume of sound sources, and placing two speakers in the simulator at the same distance~(but at different angles). Note that, since the position of the speakers is randomized, the visual signal does not provide localization cues~(\eg, via perspective). We also convert delay predictions to direction-of-arrival angles, using the known radius and microphone distance.

\mypar{Comparisons.} To provide points of comparison for this novel task, we compare our audio-visual approach with audio-only methods including GCC-PHAT. We also provide a~(oracle) baseline which selects one of the two speakers' ground-truth time delay at random, thus simulating a method that perfectly solves the localization task but which is unable to match faces to voices. 
We also consider a two-stage method that first separates the speaker's voice for each channel using VisualVoice~\cite{gao2021visualvoice}, a state-of-the-art audio-visual separation model, then applies audio-based time delay estimation methods to the separated sound. To ensure a fair comparison, we retrain the static-image based separation model~\cite{gao2021visualvoice} with input audio duration of 1.27s and 2.55s.

\begin{wraptable}{r}{0.6\textwidth}
\vspace{-2.25em}

\centering
 \caption{{\bf Quantitative results of visual-guided time delay estimation on simulated data.} We evaluate our models' ability of predicting ITD signals from mixtures with the aid of visual information.}
  \label{tab:speaker_itd}
 \resizebox{1\linewidth}{!}
 { 
  \begin{tabular}{l@{\hskip5pt}c@{\hskip5pt}c@{\hskip8pt}c@{\hskip5pt}c}
    \toprule
     Audio duration&    \multicolumn{2}{c@{\hskip8pt}}{0.96s} & \multicolumn{2}{c}{2.55s}\\
    \midrule
     Model     & RMSE  &  Err$\le 0.1\uparrow$  & RMSE  &  Err$\le 0.1\uparrow$ \\ 
    
    \midrule
    GCC-PHAT~\cite{knapp1976generalized} &   	0.503  & 56.6 & 0.504 & 56.9 \\
    Salvati et al.~\cite{salvati2021time} &    0.490  & 52.5 &  0.483 & 50.1   \\
    Random Oracle &   	   0.502  &  	56.9   & 0.502  &  	56.9 \\
    \cdashlinelr{1-5}
    Ours~-~Random   &    0.493 & 	10.0 & 0.503 &  9.76\\
    Ours~-~StereoCRW  & 	0.493  &	56.8 & 0.488 & 55.7 \\
    Ours~-~AV  &  	\textbf{0.304} & \textbf{72.5} & \textbf{0.295} & \textbf{76.1}\\
    \cmidrule{1-5}
    Sep~\cite{gao2021visualvoice}~+~GCC	& 0.361 & 	77.6 & 0.323 & 82.2  \\
    Sep~\cite{gao2021visualvoice}~+~StereoCRW 	& \textbf{0.309} & 	\textbf{82.8} & \textbf{0.281} & \textbf{85.5} \\
    \bottomrule
  \end{tabular}
  }

\vspace{-2.4em}
\end{wraptable}
\textls[-0]{
We evaluated the methods using two metrics: RMSE, and the percentage of predictions with less than 0.1 ms~(1.6 samples) of error~(Err$\le 0.1$). 
We feed models 1.0s or 2.55s audio, resulting in 512 votes.  
We show results in \tbl{tab:speaker_itd}. Our audio-visual model substantially outperforms the baselines. Interestingly, it outperforms the baseline that chooses one speaker's delay at random, an upper bound on audio-only performance. This suggests that our model successfully uses visual information. We found that the model that combines audio-visual separation with our learned audio representation performs best. Our audio-visual model (without separation) performs comparably to Sep+GCC-PHAT in regression metrics. We provide qualitative results~(for the 0.96s case) in \fig{fig:speaker_itd}.
}

\vspace{0.5mm}
\begin{wraptable}{r}{0.6\textwidth}
\vspace{-2.3em}
\captionsetup{type=figure}
    \centering
 \includegraphics[width=\linewidth]{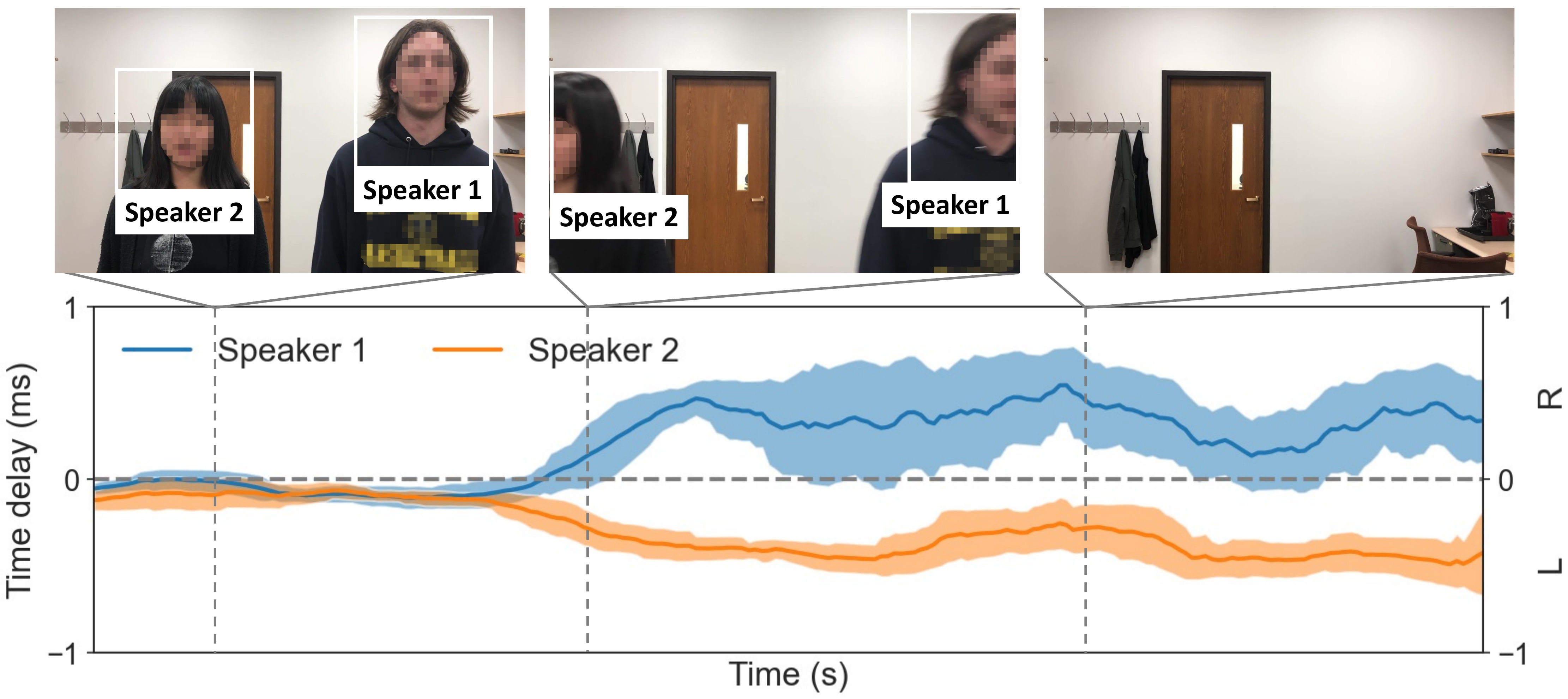}
   \caption{{\bf Visually-guided localization for a real-world scene}.  Images are blurred to preserve privacy. 
   }
   \label{fig:vitd_demo}

\vspace{-2.2em}
\end{wraptable}
\mypar{Real-world visually-guided localization.}
We perform visually-guided time delay estimation on a self-recorded video~(Fig.~\ref{fig:vitd_demo}). Two speakers talk concurrently while moving off-screen. Our model localizes each speaker in the mixture with a cropped image of their face. We show the mean and standard deviation of delay predictions in 2.0s windows.

\vspace{-1mm}
\mysection{Discussion and Limitations}
We have proposed to use {\em self-supervised time delay estimation} to localize sounds by learning interaural correspondence. We also introduced a novel {visually-guided}  localization task. Our audio models obtain performance on par with supervised methods on real-world sound, while our audio-visual model successfully localizes speakers in mixtures. We see our work opening two directions: first, integrating more visual information for multisensory localization and, second, finding finer-grained delays using recent methods from optical flow~\cite{bian2022learning,jonschkowski2020matters}. 

\mypar{Limitations.} Our audio-visual model associates the appearance of speakers with the sound of their voice. We tested on speakers that our model has been trained on, avoiding the need to generalize based solely on a person's appearance~\cite{nagrani2018seeing}. However, there is still a potential for the model to exhibit bias. Our sound-based models are trained on music, which may not be representative of all downstream tasks. We released code, data, and models on our \href{https://ificl.github.io/stereocrw}{project site}.

\mypar{Acknowledgments.} We would like to thank Xixi Hu for her valuable suggestions about the augmentation idea on random walk graphs. We thank Justin Salamon for helpful discussions and Daniele Salvati for the help on the simulator setup. We thank Zhaoying Pan and Matthew Sticha for the help recording real-world examples. We also thank Daniel Geng for his comments and feedback on the paper. This work was funded in part by DARPA Semafor and Cisco Systems. The views, opinions and/or findings expressed are those of the authors and should not be interpreted as representing the official views or policies of the Department of Defense or the U.S. Government.

\bibliographystyle{splncs04}
\bibliography{stereocrw.bib}

\supparxiv{}{
\clearpage
\appendix

\supparxiv{
\setcounter{page}{1}
}{}
\renewcommand{\thesection}{A.\arabic{section}}
\setcounter{section}{0}

\supparxiv{
\section{In-the-wild Evaluation} \label{appendix:inthewild}
\vspace{5mm}

\mypar{Video results.} 
We provide some random samples of prediction results from our in-the-wild dataset on our webpage. We also provide qualitative results for motion correlation along with our predicted time difference on our webpage. Please click \href{https://ificl.github.io/stereocrw}{here} for videos. When watching, we recommend wearing headphones, since it can be difficult to perceive stereo sound without them.

}{}

\supparxiv{
\section{Visually-guided Time delay Estimation}
\label{appendix:visualltITD}
\vspace{5mm}
\mypar{Video results.} We provide more qualitative results for the visually-guided speaker localization task along with audio on our \href{https://ificl.github.io/stereocrw}{webpage}. When watching, we recommend wearing headphones, since it can be difficult to perceive stereo sound without them. 
}{}

\section{Qualitative Results on Phone Recordings}
\label{appendix:iphone_demo}
We also ran our model on ordinary iPhone-recorded videos, exploiting the fact that portrait-mode recordings have a sufficiently large baseline for estimating time delays. We use a combination of self-collected videos and internet videos~(we used an iPhone 12 for self-recorded videos and searched Flickr for videos with tags indicating that they were recorded with an iPhone 13). We provide qualitative results in \fig{fig:iphonedemo}. %
Please see our \href{https://ificl.github.io/stereocrw}{webpage} for more video results.

\begin{figure*}[!h]
    \centering
    \vspace{-2mm}
    \includegraphics[width=\textwidth]{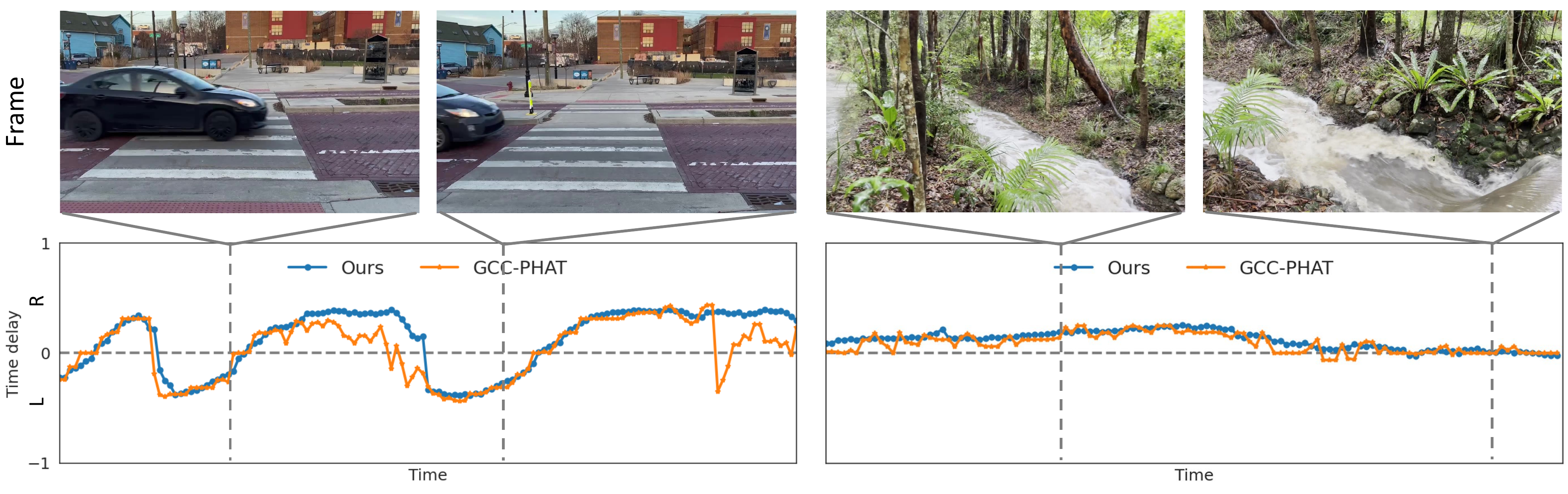}
    \caption{{\bf Qualitative results for iPhone videos.} We show time delays both for our method and for GCC-PHAT over time. The left video shot by the authors records several vehicles driving from left to right. The right video, from Flickr user \textit{Black Diamond Images}, shows a waterfall that is recorded by a moving camera. In the first frame, the waterfall is to the right. The camera then moves to face it directly. 
    }
    \label{fig:iphonedemo}
    \vspacefig
    \vspace{-4mm}
\end{figure*}

\begin{figure*}[!t]
    \centering

    \includegraphics[width=\textwidth]{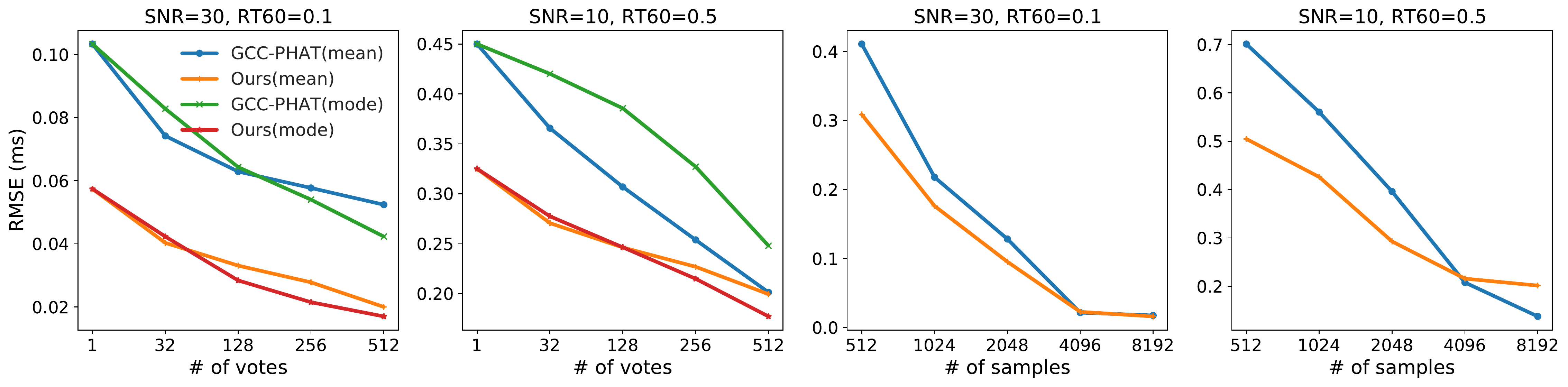}
    \begin{flushleft}
    \vspace{-3mm}
\hspace{19mm} (a) Post-Process \hspace{31mm} (b) Input duration
\vspace{-3mm}
    \end{flushleft}

    \caption{{\bf Ablation experiments on the simulated data.} (a) We evaluate with different vote numbers and post-processing methods. (b) We evaluate longer audio lengths. We note that the $x$-axis of both plots is on a log scale.  } 

    \label{fig:ablation}
    \vspacefig
\end{figure*}

\section{Ablation Study}\label{appendix:ablation}
\vspace{5mm}
\mypar{Post-processing.} 
We study the effect of the number of votes, $m$, used during post-processing for both StereoCRW and GCC-PHAT.
We evaluate 1024-sample audio using $m \in \{1,32,128,256,512\}$ for both {\em mean} and {\em mode}. In the special case $m=1$, the result is not affected by post-processing, and purely measures the quality of the representation. 
As shown in \fig{fig:ablation}(a), both methods improve with the number of votes. Our method benefits from {\em mode} post-processing, while {\em mean} works better for GCC-PHAT. In particular, we significantly outperform GCC-PHAT with $m=1$ vote, emphasizing the quality of our representation.

\begin{wraptable}{r}{0.50\textwidth}
\vspace{-2.0em}
\captionsetup{type=figure}
    \centering
\includegraphics[width=1.0\linewidth]{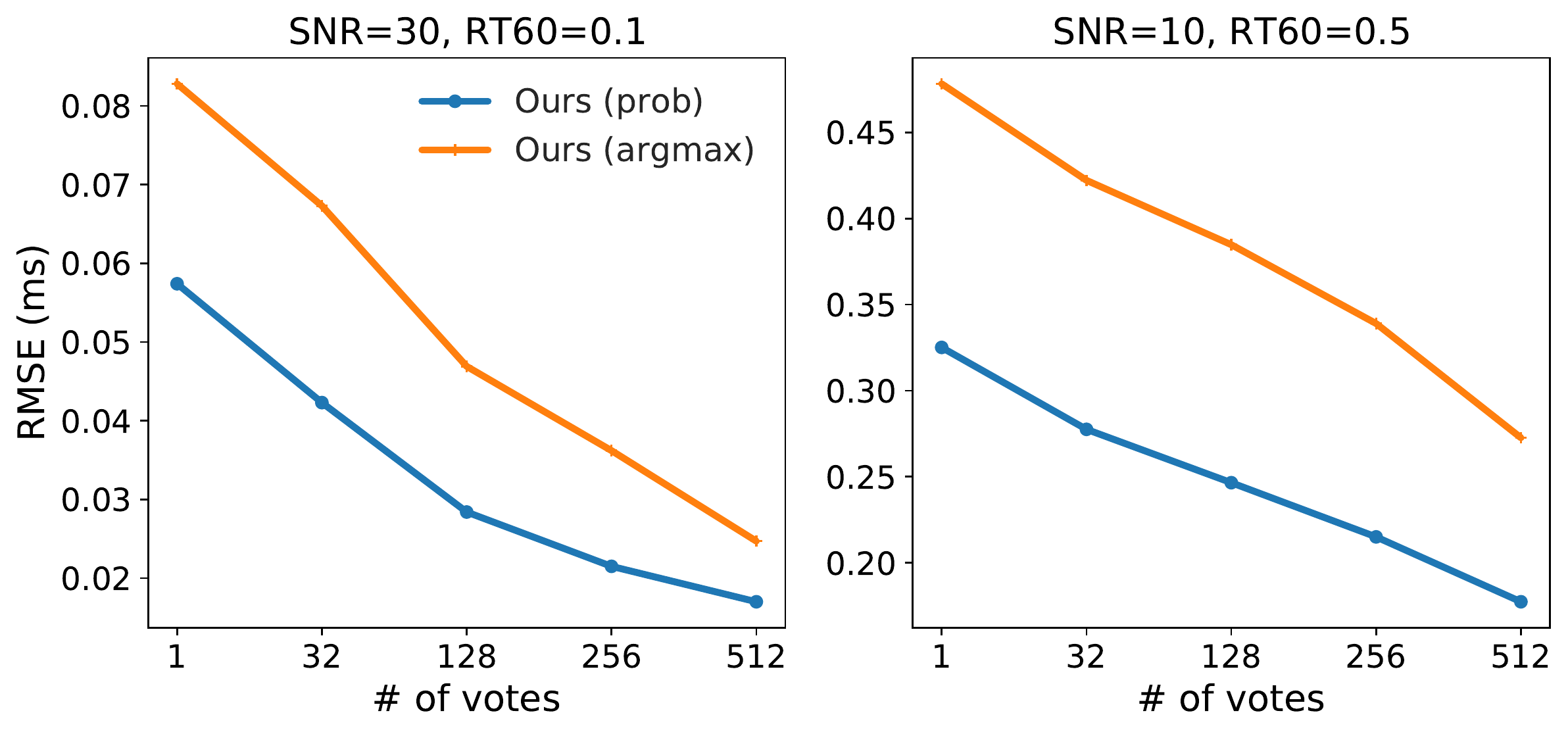}
\caption{ Probabilistic vs. Argmax.}
\label{fig:postprocess}
\vspace{-2em}
\end{wraptable}
We also design experiments to study the performance gap between using probabilistic post-processing and nearest neighbor~(argmax) for our method. For consistency, we evaluated our StereoCRW model with both types of post-processing on two simulated environments, with different vote numbers. The results are shown in \fig{fig:postprocess}. Probabilistic post-processing outperforms the nearest neighbor post-processing in the complex simulated environment. We use probabilistic post-processing in our other experiments.

\mypar{Duration.} 
We ask how our model performs when given longer audio, exploiting the fact that our embeddings use fully convolutional networks and thus can be tested on arbitrary-sized inputs~(\fig{fig:ablation}(b)). We provided our method with various input sizes~(up to $8 \times$ the training duration). For very long audio~($4 \times$ training duration), we found that our model's performance starts saturating, and that GCC-PHAT overtakes it. This may be due to the fact that the model has a fixed-size~($d = 128$) representation, while GCC-PHAT grows its representation---the waveform itself---with the input size and eventually converges to the correct solution~(a maximum likelihood estimate~\cite{knapp1976generalized,zhang2008does} in many situations).

\mypar{Simulated vs. real data.} 
We also study the data distribution gap between simulation and real-world data. We trained our best self-supervised model~(StereoCRW) and Salvati et al.~\cite{salvati2021time} on the simulated data with Free-Music-Archive clips.
We evaluated on both simulated and in-the-wild recordings (Tab.~\ref{tab:data_ablation}). As expected, our model trained on FMA-Sim obtains competitive performance, but overall does not perform as well as a model trained on real data. The supervised model improves on the in-the-wild evaluation  while the performance drops on the simulated evaluation cases when training on FMA-Sim. We also include the comparison between mode and mean post-process for Salvati et al.~\cite{salvati2021time} in the Tab.~\ref{tab:data_ablation}.

 \begin{table}[t]
\caption{\small Time delay estimation on simulated data and in-the-wild recordings. \emph{Vox-Sim} is the simulator~\cite{scheibler2018pyroomacoustics} with VoxCeleb2 clips and \emph{FMA-Sim} is the simulator with Free-Music-Archive clips.}
\label{tab:data_ablation}
\centering
\resizebox{0.9\linewidth}{!}{
\renewcommand{\arraystretch}{1.1}
\small
\begin{tabular}{l@{\hskip10pt}l@{\hskip5pt}c@{\hskip5pt}c@{\hskip5pt}c@{\hskip5pt}c@{\hskip5pt}c@{\hskip5pt}|@{\hskip5pt}c}
\hline 
  &  &  &  & &  \multicolumn{2}{c@{\hskip5pt}|@{\hskip5pt}}{Simulation} & Real-world\\
 Model & Variation & Data & Num & Aug & MAE & RMSE & Acc~($\%$)\\
\hline
\multirow{8}{*}{{\small Salvati et al.}~\cite{salvati2021time}} & {\small Mode} & Vox-Sim & 8K & & 0.146 & 0.306 & 80.0 \\
 & {\small Mean} & Vox-Sim & 8K & & {\bestcell 0.126 } & {\bestcell 0.254} & 87.7\\
 & {\small Mode} & Vox-Sim & 8K & \checkmark & 0.184 & 0.327 & 85.9\\
 & {\small Mean} & Vox-Sim & 8K & \checkmark & { 0.169} & { 0.294} & 85.8 \\
 
  & {\small Mode} & FMA-Sim & 95K & & 0.150 & 0.294  & {\bestcell 88.1}\\
 & {\small Mean} & FMA-Sim & 95K & & 0.135 & 0.256  & 88.0\\
 & {\small Mode} & FMA-Sim & 95K & \checkmark & 0.160 & 0.303  & 87.8 \\
 & {\small Mean} & FMA-Sim & 95K & \checkmark & { 0.146} & {0.267} &  87.7 \\
 \hline
StereoCRW & Mode & Vox-Sim & 8K & \checkmark &   0.193  &  0.360 & 86.4 \\
  StereoCRW & Mode & FMA-Sim & 95K & \checkmark & 0.194  &  0.341 & 87.1 \\
  StereoCRW & Mode & FMA & 95K &\checkmark &  {\bestcell 0.133} &  {\bestcell 0.259} & \textbf{87.2} \\
\hline
\end{tabular}
}
\end{table}

\section{Training with Youtube-ASMR} 
\begin{wraptable}{r}{0.52\textwidth}
\vspace{-2.2em}
\caption{{\bf Delay estimation on simulated data.} We use $\text{SNR}=10$ and $\text{RT}_{60}=0.5$s. ASMR is YouTube-ASMR~\cite{yang2020telling}. Errors in ms.  {\em Sup} refers to supervision.}
 \label{tab:asmritd_supp}
 \centering
 \resizebox{1.0\linewidth}{!}{
 \begin{tabular}{ll@{\hskip10pt}lccccc}
 \toprule
 Model & Variation & Data & Sup & Aug & MAE & RMSE \\
 \midrule
 \multirow{2}{*}{{\small Salvati et al.}~[1]}
 & {\small Mean} & Vox-Sim & \checkmark & & {\bestcell 0.126} & {\bestcell 0.254} \\
 & {\small Mean} & Vox-Sim & \checkmark & \checkmark & { 0.169} & { 0.294} \\
 \midrule
 \multirow{1}{*}{{\small GCC-PHAT}~\cite{knapp1976generalized}}
 & {\small Mean} & \multicolumn{1}{c}{--} & & &  {\bestcell 0.160} &  0.318 \\

 \midrule
 \multirow{7}{*}{Ours}
 & {\small Random} & \multicolumn{1}{c}{--} & & & 0.448 & 0.505 \\
 \cdashlinelr{2-7}
 & MonoCLR & ASMR & & & 0.425 & 0.620 \\
 & MonoCLR & ASMR & & \checkmark & 0.177 & 0.330 \\
 & ZeroNCE & ASMR & & & 0.349 & 0.468\\
 & ZeroNCE & ASMR & & \checkmark & 0.184 & {\bestcell 0.313}\\
 & StereoCRW & ASMR & & & 0.736 & 0.913\\ 
 & StereoCRW & ASMR & & \checkmark & { \bestcell0.162 } & 0.315\\ 
 \bottomrule

 \end{tabular}
}

\vspace{-3.0em}
\end{wraptable} %
We also train our models on {\bf YouTube-ASMR}, a highly diverse dataset of 30K binaural~(83 hours) internet videos~\cite{yang2020telling}. As the results shown in \tbl{tab:asmritd_supp}, our proposed ZeroNCE and StereoCRW methods outperform GCC-PHAT. Augmentation was important for YouTube-ASMR particularly, which failed to learn a good representation without it, perhaps due to the high diversity of the dataset.

\section{Simulation Setup} 
\label{appendix:sim}
We provide the details of each simulated room with its dimension and microphone positions in \tbl{tab:simulation}.

 \begin{table}[h]
 \vspace{-7mm}
\centering
 \caption{{\bf Simulation setup.} The unit is in meters.}
  \label{tab:simulation}
 \resizebox{0.75\linewidth}{!}
 { 
  \begin{tabular}{l@{\hskip10pt} c@{\hskip10pt} c @{\hskip10pt} c}
    \toprule
    & Room 1 & Room 2 & Room 3 \\
    \midrule
    Room dim~(X, Y, H)  & (7, 6, 3)  & (4, 7, 2.8)   &   (7, 7, 2.7)\\
    Left Mic position~(X, Y, H) & (3.4, 1, 1.6)  &  (0.2, 3.2, 1.7) &  (3.4, 3.1, 1.5) \\
    Right Mic position~(X, Y, H) & (3.7, 1, 1.6)  & (0.2, 3.0, 1.7) & (3.5, 2.9, 1.5)\\
    Source angle range &  $[-90^\circ, 90^\circ]$  & $[-90^\circ, 90^\circ]$  & $[-90^\circ, 90^\circ]$  \\
    Source distance range  &    $[0.5, 3.0]$ & $[0.5, 3.0]$  & $[0.5, 3.0]$ \\
    \bottomrule
  \end{tabular}
  }
  \vspace{-4mm}
  \end{table}

\section{Implementation Details }
\label{appendix:implement}
\vspace{3mm}

\mypar{Network architecture.} We use a ResNet~[38] with 9 layers as the backbone for the audio encoder. We modify the input channel number of the first convolution layer to be 2, and the output of the last fully-connected layer as 128. For a raw waveform of length $L$, we use an STFT with a window length of 256 and hop length of $\lfloor\frac{L}{128}\rfloor$ to create an input spectrogram. For the audio-visual task, we use a hop length of 160 to create an input spectrogram.

\vspace{2mm}
\mypar{Augmentations.} 
During the training, we apply the following augmentations to audio where the first three are regular augmentations applied to all the models and the last two are applied to augmented models only:
\begin{itemize}
    \item Random channel swapping: we randomly swap the left and right audio channels with a probability of 0.5.
    \item Random channel-wise scaling: we randomly re-scale each audio channel by the factor in the range of $[0.5, 1.5]$.
    \item Random shifting: for the instance discrimination model with mono audio, we randomly shift the audio for $-16$ to $16$ samples. For the audio-visual model, we apply different random shifts for each mono audio with $-24$ to $24$ samples. 
    \item Random noise: we add random Gaussian noise to audio with SNR=$[0, 30]$.
    \item Random reverberation: we add random reverberation to audio with RT$_{60}$=$[0, 0.9]$.
    \item Mixture augmentation: we randomly add another sound to the original audio. We normalize the second sound to be 10\% -- 100\% loudness of the original audio before mixing. For the audio-visual model, the second sound is normalized to be 50\%-150\% intensity level of the original one.
\end{itemize}

When computing affinity matrix $A_{21}$ for the contrastive random walk model, we do not augment $\bx_1$ with noise or mixture augmentation, so as to avoid learning unexpected matching. Similarly, for the instance discrimination model, we do not apply noise, reverberate or mixture augmentations to one of the two channels.

\vspace{1mm}
\mypar{Training details.} To accelerate the training process, we first train each model with 0.48s audio~(7680 samples) and then finetune on the input audio of 0.064s (1024 samples) using a correspondingly finer hop length.

}
\end{document}